\documentclass{interact}
\usepackage{amssymb}
\usepackage{proof}
\usepackage{hyperref}

\usepackage{pgf}
\usepackage{tikz}
	\usetikzlibrary{calc,positioning,shapes.geometric,shapes.symbols,shapes.misc, intersections, shadows, arrows, automata}

\usepackage{amsthm}
 \newtheorem{thm}{Theorem}[section]

 \theoremstyle{definition}
 \newtheorem{defn}[thm]{Definition}
 \theoremstyle{remark}

\usepackage{pgf}
\usepackage{xxcolor}

\usepackage{natbib}

\usepackage{mycls}

\def\qed{\relax
   \ifmmode
    ~\hfill\Box
   \else
    \unskip\nobreak ~\hfill$\square$%
   \fi \par}

\begin{document}
	\title{Automated Reasoning in Normative Detachment Structures with Ideal Conditions}
\author{Tomer Libal and Matteo Pascucci}
	\date{}
	\maketitle

\begin{abstract}
Systems of deontic logic suffer either from being too expressive and therefore hard to mechanize, or from being too simple to capture relevant aspects of normative reasoning.
In this article we look for a suitable way in between: the automation of a simple logic of normative ideality and sub-ideality that is not affected by many deontic paradoxes and that is expressive enough to capture contrary-to-duty reasoning.
We show that this logic is very useful to reason on normative scenarios from which one can extract a certain kind of argumentative structure, called a Normative Detachment Structure with Ideal Conditions. The theoretical analysis of the logic is accompanied by examples of automated reasoning on a concrete legal text.
\end{abstract}


\section{Introduction}\label{introduction}

In the last decades, computer systems have played an  important role in assisting people in a wide range of tasks, from searching over data to decision-making, and their use is required in an increasing number of fields. 
One of these fields is \textit{legal reasoning}. New court cases and legislations are accumulated every day. In addition, international
organizations like the European Union are constantly aiming at combining and integrating separate legal systems (\citealt{court}). However, the automation of legal reasoning is still underdeveloped.
In recent years, some automatic procedures have been developed in order to deal with courtroom management\footnote{\url{http://softpert.com/legal/court-management/winjuris}} or
legal language processing and management (\citealt{eunomos}); moreover, several expert systems based on cases or rules have been introduced (\citealt{splitup,offender}); finally, some logical systems for automatic reasoning over sets of norms such as the HIPAA and GLBA
privacy laws have been presented (\citealt{hipaa}).

In this article we focus on the automation of logical systems used to reason on legal texts. One of the main issues in this area of research is finding a good balance between the expressivity of logical languages and the efficiency of theorem provers. Indeed, 
the variety of concepts employed in legal reasoning suggests the choice of very complex formal languages and systems; however, this entails that
existing and efficient tools for automated reasoning cannot  be straightforwardly exploited. Therefore, one has to look for some logical framework that is expressive enough to capture the fundamental aspects of the problems of legal reasoning to be addressed and that behaves well from the point of view of automation.

A very simple logical system dealing with normative concepts is \textbf{SDL} (\textit{Standard Deontic Logic}). This system can be easily encoded in theorem provers but is affected by seemingly unsurmountable difficulties in representing very common normative scenarios, such as those in which a contrary-to-duty norm applies (\citealt{navarro}). On the other hand, there are extensions of \textbf{SDL} that are still fairly simple but do not share most of its weaknesses.
%
One of such extensions is proposed by \cite{J&P85} and called \textbf{DL}. The latter system aims at a rigorous representation of the difference between normative ideality and sub-ideality, a difference that is not expressible in \textbf{SDL}; here we will consider a variation of \textbf{DL} suggested by \cite{deBoer&al.12}, which will be called \textbf{DL}$^*$. We will show that \textbf{DL}$^*$ can be suitably exploited to reason on normative scenarios from which one can extract an argumentative structure involving a list of ideal normative statements, a list of normative conditionals, a list of factual relations among the various statements and some actual circumstances. Such structure will be called a \textit{Normative Detachment Structure with Ideal Conditions}.
Specifically, we will see
how, extracting this structure from a legal text, one can use \textbf{DL}$^*$ to formulate normatively relevant queries about the text; moreover, we will show that these queries can be answered by a theorem prover.


The closest work to ours is \cite{hol}, where the authors translate the language of deontic logics significantly more complex than \textbf{DL}$^*$ into higher-order languages 
and then encode the latter into automatic theorem provers such as Isabelle/HOL (\citealt{isabelle})
and LEO-3 (\citealt{leo3}). 
The major difference with our approach turns out to be a matter of complexity in derivability-checking.

The article is structured as follows. Section \ref{Sec 2} is devoted to a thorough presentation of logics of normative ideality and sub-ideality, including some detailed motivations for the choice of \textbf{DL}$^*$.
In the same section we also present the argumentative structure of the normative scenarios we want to deal with.
In section \ref{Sec 3}
we provide an example of a legal text and formalize its core sentences within the language of our logical system. In section \ref{sec:is} we represent some normatively relevant queries as problems of derivability of formulas in \textbf{DL}$^*$. Furthermore, we introduce a program which
can answer similar queries in a fully-automated way.
Finally, we conclude our work with some theoretical reflection on the representation of contrary-to-duty scenarios proposed in the article.

\section{Logics of normative ideality and sub-ideality}\label{Sec 2}
Works on formal approaches to normative reasoning usually start with a radical criticism of the so-called system of Standard Deontic Logic (\textbf{SDL}), which is the weakest normal deontic system closed under the schema $OA\rightarrow \neg O\neg A$, namely the deontic version of the alethic system {\bf{KD}}. Indeed, authors usually point out a list of theorems of \textbf{SDL} that are associated with paradoxes of deontic reasoning. For instance, the provable schema $OA \rightarrow O(A\vee B)$ gives rise to \textit{Ross's paradox} when the formulas $A$ and $B$ represent, respectively, the propositions expressed by sentences like `Mark posts the letter' and `Mark burns the letter': the inference from `Mark ought to post the letter' to `Mark ought to post the letter or burn it' is at least difficult to justify. While these paradoxes are sometimes due to ambiguities in the natural language sentences to be formalized, there is a more important flaw of \textbf{SDL} which concerns the formalization of \textit{contrary-to-duty obligations}. This problem is exemplified by \textit{Chisholm's paradox}.\footnote{For an extended discussion of the paradox and a presentation of basic aspects of deontic logic, we refer the reader to \cite{Carmo&Jones02}.} Consider the following set of sentences:
\begin{itemize}
\item[(1)] it ought to be that Jane helps her neighbors;
\item[(2)] it ought to be that if Jane helps her neighbors, she tells them that she is coming;
\item[(3)] if Jane does not help her neighbors, then she ought not to tell them that she is coming;
\item[(4)] Jane does not help her neighbors.
\end{itemize}
Under any plausible formalization in the language of \textbf{SDL}, these sentences turn out to be either inconsistent or not logically independent and both  outcomes are clearly undesirable.

On the other hand, the reasons of the aforementioned drawbacks are often overlooked. As \cite{J&P85} claim, many issues arise from the interpretation of the operator $O$. Indeed, the semantic intuition associated with a formula of kind $OA$ in \textbf{SDL} is that $A$ is true in all \textit{normatively ideal circumstances} (or worlds), namely in all those circumstances in which \textit{every} prescription is observed. However, the majority of the sentences which describe propositions true in all normatively ideal circumstances are not normatively relevant, such as the sentence  `it either rains or does not rain'. Thus, in order to formally capture the meaning of `ought'-sentences, one has to take into account \textit{some criterion of normative relevance for sentences}. The proposal made by \cite{J&P85} consists in requiring an `ought'-sentence to describe a proposition which not only holds in all normatively ideal circumstances, but also fails in some \textit{normatively sub-ideal circumstance}, namely in some circumstance in which \textit{not every} prescription is observed. For instance, if Jane ought to help her neighbors, then one can say that this happens to be the case in all normatively ideal scenarios, but fails to be the case in some normatively sub-ideal scenario.

In order to distinguish between normative ideality and sub-ideality, Jones and P\"{o}rn extend the language of \textbf{SDL} with an operator $O'$ such that the formula $O'A$ means that $A$ is true in all normatively sub-ideal worlds. Then, they propose the following formalization of `ought'-sentences: $Ought (A) =_{def} OA \wedge \neg O'A$. The system obtained with the addition of $O'$ is called \textbf{DL} and represents a bimodal version of \textbf{SDL} supplemented with the axiom-schema $(OA\wedge O'A)\rightarrow A$. Actually the latter schema is not explicitly mentioned by the authors, but is valid in the intended semantics, as it can be easily shown (see also \citealt{deBoer&al.12}).  Frames to interpret \textbf{DL} are structures of kind $\frak{F}=\langle W, R_O, R_{O'}\rangle$, where $W$ is a domain of worlds and $R_O$ and $R_{O'}$ are binary relations over $W$ satisfying the following properties:
\begin{itemize}
\item[(I)] for all $w\in W$, there are $v,u\in W$ s.t. $wR_O v$ and $wR_{O'}u$;
\item[(II)] for all $w\in W$,  $R_O(w)\cap R_{O'}(w)=\emptyset$;
\item[(III)] for all $w\in W$, either $w\in R_O(w)$ or $w\in R_{O'}(w)$.
\end{itemize}
Property (I) can be captured already by the axiomatic basis of bimodal \textbf{SDL}, property (II) requires further discussion that will be provided below and property (III) can be captured only if one extends the axiomatic basis of bimodal \textbf{SDL} with a schema such as $(OA\wedge O'A)\rightarrow A$. Thus, if one wants \textbf{DL} to be the logic characterized by the class of frames satisfying properties (I), (II) and (III), as it seems to be suggested by Jones and P\"{o}rn, then \textbf{DL} has to be a proper extension of bimodal \textbf{SDL}.

Despite its broader expressive power, \textbf{DL} still encounters some obstacles in dealing with contrary-to-duty obligations. Indeed, as observed in \cite{Prakken&Sergot96}, it gives  rise to a `pragmatic oddity' when the sentences (1)-(4) above are formalized as suggested by Jones and P\"{o}rn, namely (taking $P$ to be `Jane helps her neighbors' and $Q$ to be `Jane tells her neighbors that she is coming'):
\begin{itemize}
\item[(1a)] $Ought(P)$;
\item[(2a)] $O(P\rightarrow Ought(Q)) \wedge O'(P\rightarrow Ought(Q))$;
\item[(3a)] $O(\neg P\rightarrow Ought(\neg Q)) \wedge O'(\neg P\rightarrow Ought(\neg Q))$;
\item[(4a)] $\neg P$.
\end{itemize}
The oddity is hidden in the fact that in \textbf{DL} (1a)-(4a) entail both $O P$ and $O\neg Q$, which means that in all normatively ideal worlds Jane helps her neighbors without telling them that she is coming. Furthermore, \cite{Hansson89} shows that certain instances of paradoxes of deontic reasoning still hold in \textbf{DL}, such as the following version of Ross's paradox: if Mark neither posts the letter nor burns it, while he ought to post it, then he ought to post it or burn it. Indeed, the schema $(\neg A\wedge \neg B)\rightarrow (Ought(A)\rightarrow Ought(A\vee B))$ is provable in {\bf{DL}}.

The latter problem finds a remedy in \cite{deBoer&al.12}, where the authors propose to replace the operator $Ought$ with an operator $Ought^*$ such that $Ought^*(A)=_{def} OA\wedge O'\neg A$; in this way the problematic schemata mentioned by Hansson are no longer provable.
However, there is a fundamental aspect of $Ought^*$ which requires further analysis. Since $Ought^*(A)$ is true at a world only if $O'\neg A$ is true there, then the meaning of $O'$ proposed by Jones and P\"{o}rn needs to be revised. Indeed, otherwise one would have that \textit{every} prescription is violated in all sub-ideal worlds, which is implausible, since a world can be classified as normatively sub-ideal even if \textit{some but not all} prescriptions are violated there. Thus, in order to exploit the operator $Ought^*$ it is better to take $O'A$ as meaning that $A$ holds in all \textit{normatively awful} worlds. Notice that this reading allows one to get rid of the schema $(OA\wedge O'A)\rightarrow A$, as well as of the frame condition (III) associated with it, since the current world might turn out to be neither normatively ideal nor normatively awful (from the perspective of the norms currently in effect). From the point of view of sub-ideality this change is not dramatic: a world is still  classified as sub-ideal if and only if it is not an ideal one.

Hereafter we will denote by \textbf{DL}$^*$ the logic resulting from \textbf{DL} by removing the axiom $(OA\wedge O'A)\rightarrow A$ and adding the definition of $Ought^*$.
A crucial issue is whether \textbf{DL}$^*$ coincides with a bimodal version of \textbf{SDL}. This is actually the case, since \cite{deBoer&al.12} show (Lemma 3.5) that every model over a frame which violates property (II) can be transformed into an equivalent model over a frame satisfying property (II); thus, since property (I) can be captured already by the axioms of bimodal \textbf{SDL}, there is no need to add further postulates to get a characterization result for \textbf{DL}$^*$ with respect to its intended class of frames, namely the class of frames satisfying properties (I) and (II).

It is important to remark that the `pragmatic oddity' may affect the formalization of Chisholm's example even if one uses $Ought^*$ in place of $Ought$. For instance, \cite{deBoer&al.12} argue that, in the absence of property (III), the formalization of sentences (2) and (3) needs the following revision in order to allow for the detachment of $Ought^* \neg Q$, which is an intended consequence of the scenario (Jane ought not to tell her neighbors that she is coming, since she decided not to help them):
\begin{itemize}
\item[(2a')] $(P\rightarrow Ought^*(Q)) \wedge O(P\rightarrow Ought^*(Q)) \wedge O'(P\rightarrow Ought^*(Q))$;
\item[(3a')] $(\neg P\rightarrow Ought^*(\neg Q)) \wedge O(\neg P\rightarrow Ought^*(\neg Q)) \wedge O'(\neg P\rightarrow Ought^*(\neg Q))$.
\end{itemize}
However, from (1a), (2a'), (3a') and (4a) one still gets both $OP$ and $O\neg Q$ as consequences and so an implausible description of what is the case in all normatively ideal worlds (Jane helps her neighbors without telling them).

A solution to the pragmatic oddity can be formulated by combining the use of $Ought^*$ (which is beneficial anyway, since it allows one to get rid of Hansson's version of deontic paradoxes) and an intuition discussed in \cite{J&P85}, according to which the first sentence of a Chisholm-like scenario  expresses an obligation which holds in ideal circumstances, hereafter simply called an \textit{ideal obligation}. For instance, one can imagine that in ideal circumstances Jane ought to help her neighbors, while in the present circumstance something prevents her from doing that; thus, she does not have an \textit{actual obligation} to help her neighbors.\footnote{It has to be remarked that, according to this suggestion by Jones and P\"{o}rn, a world $w$ is normatively ideal with respect to a world $w'$  if an only if:
\begin{itemize}
\item all prescriptions that are \textit{actual} in $w'$ are observed in $w$;
\item all prescriptions that are ideal in $w'$ apply to $w$.
\end{itemize}
However, it would be more appropriate to say that there are \textit{two levels of ideality} from the perspective of a world $w$: the first level is that of any world $w'$ s.t. $w R_O w'$, which represents \textit{ideality with respect to what is actually prescribed} in $w$; the second level is that of any world $w''$ s.t. $w'R_Ow''$ and $wR_Ow'$, which represents \textit{ideality in a strict sense} (everything which is ideally prescribed in $w$ is observed in $w''$). Furthermore, notice that, as claimed by \cite{Carmo&Jones02}, the distinction between ideal prescriptions and actual prescriptions does not coincide with the distinction between \textit{prima facie} prescriptions and \textit{all-things-considered} prescriptions, which is often invoked in defeasible reasoning. Indeed, among the set of \textit{prima facie} prescriptions one can have both ideal prescriptions and non-ideal ones. The notion of ideal prescription makes reference to an implicit normative standard. In Chisholm's example, Jane has the \textit{prima facie} obligation of helping her neighbors and the \textit{prima facie} obligation of telling them that she is not coming (as soon as she decides not to help them); however, the normative standard applies only to the first prescription, since in ideal circumstances she ought to help her neighbors.}
The result is the following formalization of sentence (1):
\begin{itemize}
\item[(1a')] $O (Ought^* (P))$.
\end{itemize}
Moreover, we propose here three modifications of (2a') and (3a'). First, since the reading of $O'$ has to be changed from `in all normatively sub-ideal worlds' to `in all normatively awful worlds' and normatively awful worlds cannot be expected to verify conditional obligations, then the conjuncts $O'(P\rightarrow Ought^*(Q))$ and $O'(\neg P\rightarrow Ought^*(\neg Q))$ can be dropped from (2a') and (3a'). Second, we remove also the conjuncts $O(P\rightarrow Ought^*(Q))$ and $O(\neg P\rightarrow Ought^*(\neg Q))$, since in more complex scenarios   conjuncts of this kind could allow one to infer an ideal obligation, $O(Ought^*(B))$, from two premises $Ought^*(A)$ and $O(A \rightarrow Ought^*(B))$; indeed, $Ought^*(A)$ entails $OA$ and this, together with  $O(A \rightarrow Ought^*(B))$, entails $O(Ought^*(B))$. The point is that it is clearly not acceptable to infer an ideal obligation from an actual one.
Third, we want the formal representation of a conditional obligation to allow for the construction of a chain of statements that provide a full description of what ideally ought to be the case. For instance, we know that Jane \textit{ideally} ought help her neighbors and that the fact that she helps her neighbors entails that she ought to tell them that she is coming; from this one would like to infer that Jane \textit{ideally} ought to tell her neighbors that she is coming. In order to get this result without affecting a uniform rendering of conditional obligations, we add to (2a') the conjunct $O(Ought^*(P)\rightarrow Ought^*(Q))$ and to (3a') the conjunct $O(Ought^*(\neg P)\rightarrow Ought^*(\neg Q))$. Notice that in the former case one gets $O(Ought^*(Q))$ from (1a') and the schema $O(A\rightarrow B)\rightarrow (OA\rightarrow OB)$, whereas in the latter case no ideal obligation can be detached, since the antecedent itself is not an ideal obligation. The result of all modifications is the following rendering of sentences (2) and (3):
\begin{itemize}
\item[(2a'')] $(P\rightarrow Ought^*(Q))\wedge O(Ought^*(P)\rightarrow Ought^*(Q))$;
\item[(3a'')] $(\neg P\rightarrow Ought^*(\neg Q))\wedge O(Ought^*(\neg P)\rightarrow Ought^*(\neg Q))$.
\end{itemize}
The formula $OP$ is not derivable from (1a'),(2a''), (3a'') and (4a), so the pragmatic oddity no longer arises;\footnote{This can be justified in terms of the two levels of ideality described in the previous footnote.} furthermore, the four premises are still logically independent and consistent. The key aspect of this solution is that obligations with respect to ideal worlds (ideal obligations), such as $O(Ought^*(P))$, are kept distinct from obligations with respect to the actual world (actual obligations), such as $Ought^*(\neg Q)$. 

We can generalize this approach to scenarios that are more complex than Chrisholm's paradox and that include:
\begin{itemize}
\item[(i)] a list of ideal normative statements;
\item[(ii)] a list of normative conditionals;
\item[(iii)] some factual relations among the statements in (i) and (ii);
\item[(iv)] some circumstances which trigger the antecedents of some conditionals in (ii).
\end{itemize}
We can call this argumentative structure a  \textit{Normative Detachment Structure with Ideal Conditions} (hereafter $NDSIC$). Let us consider the case in which all normative statements involved in an $NDSIC$ are obligations, then this structure can be more precisely described as follows (for some $n,k\geq 0$ and some $m\geq 1$):
\begin{itemize}
\item[($Cid_1$)] $A^*_1$ \textit{ideally} ought to be the case;
\item[] ...
\item[($Cid_n$)] $A^*_n$ \textit{ideally} ought to be the case;
\item[($Ccon_1$)] if $A_1$ then $B_1$ ought to be the case;
\item[] ... 
\item[($Ccon_m$)] if $A_m$ then $B_n$ ought to be the case;
\item[($Crel_1$)] some relation $Rel_1$ among the statements involved in ($Cid_1$)-($Ccon_m$) holds;
\item[] ...
\item[($Crel_k$)] some relation $Rel_k$ among the statements involved in ($Cid_1$)-($Ccon_m$) holds;
\item[($Cant$)] the antecedents of some conditionals in ($Ccon_1$)-($Ccon_m$) hold.
\end{itemize}
In a structure of this kind one can detach both actual obligations (triggered by ($Cant$)) and ideal obligations (triggered by ($Cid_1$)-($Cid_n$)) from the list of conditionals.

For instance, take the following example of an $NDSIC$, extracted from the travel guidelines of passengers of the Sociedade de Transportes Colectivos do Porto:\footnote{\url{http://www.stcp.pt/en/travel/how-to-travel/}}
\begin{itemize}
\item[(1b)] ideally it ought to be the case that passengers have their ticket ready in advance before boarding;
\item[(2b)] if passengers have their ticket ready in advance, then they ought to validate it immediately after boarding;
\item[(3b)] if passengers do not have their ticket ready in advance, then they ought to buy it immediately after boarding;
\item[(4b)] if passengers ought to buy their ticket immediately after boarding, then they ought to pay with the exact amount of money.
\end{itemize}
Furthermore, assume that
\begin{itemize}
\item[(5b)] passengers do not have their ticket ready in advance.
\end{itemize}
From this scenario one can infer that passengers actually ought to buy their ticket immediately after boarding and (as a consequence) actually ought to pay with the exact amount of money. However, one cannot infer that passengers \textit{actually} ought to have their ticket ready in advance. Here (1b) is a clause of kind ($Cid$) and (2b)-(4b) are clauses of kind ($Ccon$).

Notice that an $NDSIC$ is not a contrary-to-duty structure on its own; a contrary-to-duty feature emerges when some clause of kind ($Cid$), which describes an ideal normative statement, conflicts with the clause  ($Cant$), which describes what actually is the case and triggers some normative statement involved in the list of conditionals; in this situation, some of the normative statements triggered by ($Cant$) are contrary-to-duty ones. On the other hand, if there is no conflict between the ideal and the actual, then an $NDSIC$ simply represents an argumentative structure in which normative statements can be detached from conditionals given the actual circumstances and the clauses ($Cid_1$)-($Cid_n$). 

We conclude this section with another important comment on the logic \textbf{DL}$^*$: the definition of a plausible operator of permission. One cannot simply take the dual of $Ought^*$, since $\neg Ought^*(\neg A)$ means $\neg(O \neg A \wedge O'A)$, namely $P A \vee P'\neg A$, which is too weak to express permission. Also in this case, one can borrow a solution from \cite{J&P85} and have $Perm^*(A)=_{def} PA \wedge P'\neg A$. According to such definition and the revised reading of $O'$, $A$ is permitted iff there is a normatively ideal world in which it holds and a normatively awful world in which it does not hold. The first conjunct witnesses that $A$ is compatible with everything which should be the case; the second conjunct that $A$ is not trivially true in all possible scenarios.


\section{Reasoning in legal texts}\label{Sec 3}

When attempting to implement a system capable of reasoning about legal texts, different types of reasoning emerge; \cite{js92} discuss two of them, \textit{definitional}  (or qualitative) reasoning and \textit{normative} reasoning. Definitional reasoning
specifies the conditions ``under which some entity $x$ counts as an entity of a particular type $F$''; for instance, if $x$ was born in the UK, then $x$ is a British citizen. Normative reasoning concerns the distinction between what is the case in normatively ideal circumstances and what is the case in actual circumstances.
While both kinds of reasoning occur regularly in legal texts, we will focus on the second one, which can be naturally handled within the logical framework provided in the previous section. 

In order to see how the logic \textbf{DL}$^*$ can be used to formally represent normative reasoning, we can borrow an example of a legal text from \cite{js92}, \textit{The United Nations Convention on Contracts for the International Sale of Goods}. In section \ref{convention} we provide some articles from the 2010 version of the Convention.\footnote{\url{https://www.uncitral.org/pdf/english/texts/sales/cisg/V1056997-CISG-e-book.pdf}}
We use \textbf{DL}$^*$ to represent some \textit{normatively relevant questions} related to this text; more specifically, we consider a situation in which an international transaction has just been concluded and either the buyer or the seller wants to navigate through the directives of the Convention to understand which are the normative consequences of the transaction. 

\subsection{The United Nations Convention on Contracts for the International Sale of Goods}\label{convention}

\textit{The United Nations Convention on Contracts for the International Sale of Goods} represents a common structure in legal texts.
Here we will show how an $NDSIC$ can be extracted from a sample of articles in the Convention. These articles describe the duties of the seller and the rights and duties of the buyer in case of a transaction.
\\\\

{\bf Article 30}\\
The seller must deliver the goods, hand over any documents relating to
them and transfer the property in the goods, as required by the contract and
this Convention.
\\

{\bf Article 31}\\
If the seller is not bound to deliver the goods at any other particular
place, his obligation to deliver consists:
\begin{enumerate}
  \item[(a)] if the contract of sale involves carriage of the goods -- in handing
the goods over to the first carrier for transmission to the buyer;
\item[(b)] if, in cases not within the preceding subparagraph, the contract relates to specific goods, or unidentified goods to be drawn from a specific stock or to be manufactured or produced, and at the time of the conclusion of the contract the parties knew that the goods were at, or were to be manufactured or produced at, a particular place -- in placing the goods at the buyer's disposal at that place;
  \item[(c)] in other cases -- in placing the goods at the buyer's disposal at the
place where the seller had his place of business at the time of the conclusion
of the contract.
\end{enumerate}

{\bf Article 32}
\begin{enumerate}
  \item If the seller, in accordance with the contract or this Convention,
    hands the goods over to a carrier and if the goods are not clearly identified
    to the contract by markings on the goods, by shipping documents or
    otherwise, the seller must give the buyer notice of the consignment ­specifying
    the goods.
  \item If the seller is bound to arrange for carriage of the goods, he must
    make such contracts as are necessary for carriage to the place fixed by means
    of transportation appropriate in the circumstances and according to the usual
    terms for such transportation.
  \item If the seller is not bound to effect insurance in respect of the
    carriage of the goods, he must, at the buyer’s request, provide him with all
    available information necessary to enable him to effect such insurance.
\end{enumerate}

{\bf Article 45}
\begin{enumerate}
  \item If the seller fails to perform any of his obligations under the
    contract or this Convention, the buyer may:
    \begin{enumerate}
      \item exercise the rights provided in articles 46 to 52;
      \item  claim damages as provided in articles 74 to 77.
    \end{enumerate}
  \item The buyer is not deprived of any right he may have to claim
    damages by exercising his right to other remedies.
  \item No period of grace may be granted to the seller by a court or arbitral
    tribunal when the buyer resorts to a remedy for breach of contract.
\end{enumerate}

{\bf Article 53}\\
The buyer must pay the price for the goods and take delivery of them
as required by the contract and this Convention.
\\\\

Let us first provide a brief and informal reconstruction of the relations among the various Articles in this part of the Convention.
Article 30 describes what ideally ought to be the case when a transaction has been made: the seller is ideally required to deliver the goods, hand over the documents and transfer the property in the goods.
Article 31 specifies some conditional obligations depending on the ideal duties of the seller: if the seller is committed to carriage of the goods, then he/she must hand the goods over to the first carrier; in other cases, he/she needs to place the goods at the buyer's disposal at a specified place.
Article 32 includes further conditional obligations triggered by the situation in which the seller hands the goods over to the first carrier available (i.e., by one of the scenarios illustrated in Article 31).
Article 45 introduces a contrary-to-duty feature: if the seller does not fulfill his/her duties described in the previous Articles, then the buyer can exercise some right and claim damage.
Article 53 describes what ideally ought to be the case when a transaction has been made: the buyer is ideally required to pay and take delivery of the goods.

\subsection{A formal representation of the Convention}\label{sec:form}

There are many ways in which one can move from the natural language sentences of a legal text, such as the Convention, to formulas of a logical language. The degree of success of the proposed formalization of a text can be evaluated in terms of the set of inferences allowed by it, which should correspond to natural language inferences supported by one's intuitions.

\begin{defn}[Language of the Convention]
  \label{df:syntax}
We start by codifying some statements from the Convention as propositional symbols (constants):
\begin{itemize}
  \item $D$ -  the seller delivers the goods, hands over the documents and transfers the property according to the procedure described in the contract;
  \item $D0.1$ - the contract requires the seller to take care of the carriage of goods;
  \item $D0.2$ - the contract relates to goods to be produced at a particular place;
  \item $D1$ - the goods are handed over to the first carrier;
  \item $D1.1$ the goods are clearly identified by markings, shipping documents, etc.;
  \item $D1.2$ the seller notifies the buyer of the consignment;
    \item $D1.3$ the seller makes contracts necessary for carriage;
      \item $D1.4$ the seller is bound to effect insurance in respect of the carriage;
        \item $D1.5$ the seller provides the buyer information to effect the insurance for carriage;
  \item $D2$ - the goods are disposed at the place of production;
  \item $D3$ - the goods are disposed at the business address of the seller;
  \item $E1$ - the buyer exercises rights;
  \item $E2$ - the buyer claims damages;
  \item $G$ - the buyer takes delivery of the goods;
  \item $P$ - the buyer pays for the goods.
\end{itemize}
Then, we introduce deontic operators and normative conditionals that can be expressed in \textbf{DL}$^*$; we employ here a different and simplified notation which points out their reading in a more explicit way. Let $A$ and $B$ be propositional formulas, then:
\begin{itemize}
  \item $Id(A)$ - $A$ holds in all normatively ideal circumstances (this corresponds to the formula $OA$ in \textbf{DL}$^*$);
  \item $Aw(A)$ - $A$ holds in all normatively awful circumstances (this corresponds to the formula $O'A$ in \textbf{DL}$^*$);
  \item $Ob(A)$ - $A$ ought to be the case (this corresponds to the formula $Ought^*(A)$ in \textbf{DL}$^*$);
  \item $Pm(A)$ - $A$ can be the case (this corresponds to the formula $Perm^*$ in \textbf{DL}$^*$)
  \item $A\Rightarrow_{Ob} B$ - $B$ is an obligation triggered by condition $A$ (this corresponds to the formula $(A\rightarrow Ought^*(B))\wedge O(Ought^*(A)\rightarrow Ought^*(B))$ in \textbf{DL}$^*$);
  \item $A\Rightarrow_{Pm} B$ - $B$ is a permission triggered by condition $A$ (this corresponds to the formula $(A\rightarrow Perm^*(B))\wedge O(Perm^*(A)\rightarrow Perm^*(B))$ in \textbf{DL}$^*$).
\end{itemize}
\end{defn}


\begin{defn}[The Convention]
  \label{df:un}
  The following is the formal description of a relevant set of statements from the Convention.
  \begin{enumerate}
    \item $Id(Ob(D))$;
    \item $Id(Ob(P))$;
    \item ${D0.1} \Rightarrow_{Ob} {D1}$;
    \item $D0.2 \Rightarrow_{Ob} D2$;
    \item $(\neg D0.1 \wedge \neg D0.2)\Rightarrow_{Ob} D3$;
    \item ${D} \Rightarrow_{Ob} {G}$;
    \item ${D1} \Rightarrow_{Ob} {D1.3}$;
    \item $({D1\wedge\neg D1.1}) \Rightarrow_{Ob} {D1.2}$;
    \item $({D1\wedge\neg D1.2}) \Rightarrow_{Ob} {D1.1}$;
    \item $({D1\wedge\neg D1.4}) \Rightarrow_{Ob} {D1.5}$;
    \item $({D1\wedge\neg D1.5}) \Rightarrow_{Ob} {D1.4}$;
    \item ${\neg D}\Rightarrow_{Pm}{E1}$;
    \item ${\neg D}\Rightarrow_{Pm}{E2}$;
    \item $D0.1 \rightarrow \neg D0.2$;
    \item $D1\rightarrow \neg D2$;
    \item $D1 \rightarrow \neg D3$;
    \item $D2 \rightarrow \neg D3$;
    \item $D\rightarrow [(D0.1\rightarrow (D1 \wedge (D1.1 \equiv \neg D1.2)\wedge D1.3 \wedge (D1.4\equiv \neg D1.5)))\wedge(D0.2\rightarrow D2)\wedge ((\neg D0.1\wedge\neg D0.2)\rightarrow D3)]$.
  \end{enumerate}
\end{defn}

Let $UN$ bet the set of formulas (1)-(18) above.
By extending $UN$ with a set of actual circumstances playing the role of ($Cant$), one gets an $NDSIC$, where (1)-(2) stand for clauses of kind ($Cid$), (3)-(13) for clauses of kind ($Ccon$) and (14)-(18) for clauses of kind ($Crel$).
We will next consider some normatively relevant queries concerning this scenario, that is queries whose answer can help the buyer and the seller in understanding the normative consequences of the Convention.

\section{Normative queries about the Convention}
\label{sec:is}

In this section we provide some examples of queries concerning the Convention that can be formally represented within the language of \textbf{DL}$^*$, in accordance with Definition \ref{df:syntax} and Definition \ref{df:un}. These queries might be formulated by a buyer or a seller who has just concluded a transaction and wants to explore its normative consequences.

The first group of problems (Queries 1-2) concerns the compatibility of a given scenario with the Convention. We can assume that the Convention is violated whenever something that is ideally obligatory does not hold.
We use $\bigwedge UN$ as an abbreviation for the conjunction of formulas (1)-(18).

\paragraph*{Query 1.}
Is there any violation of the Convention if the contract requires the seller to take care of the carriage of the goods and they are placed at the seller's address of business?
\\\\
This problem can be expressed as a question about the derivability of the following conditional in \textbf{DL}$^*$: $(\bigwedge UN \wedge D0.1 \wedge D3)\rightarrow(\neg D\vee\neg P\vee\neg G)$. If this conditional is derivable in \textbf{DL}$^*$, then the situation described in Query 1 actually represents a violation of the Convention.

\paragraph*{Query 2.}
Is there any violation of the Convention if the contract neither requires the seller to take care of the carriage of the goods nor refers to goods that have to be produced at a particular place and the seller does not dispose the goods at his/her place of business?
\\\\
This problem can be expressed as a question about the derivability of the following conditional in \textbf{DL}$^*$: $(\bigwedge UN \wedge \neg D0.1 \wedge \neg D0.2 \wedge \neg D3)\rightarrow(\neg D\vee\neg P\vee\neg G)$. If this conditional is derivable in \textbf{DL}$^*$, then the situation described in Query 2 actually represents a violation of the Convention.
\\

The second group of problems (Queries 3-6) concerns the detachment of normative statements from given scenarios. The last three problems in this group (Queries 4-6) make reference to contrary-to-duty scenarios.

\paragraph*{Query 3.}
In case the contract requires the seller to take care of the carriage of the goods, does the seller have to notify the buyer of consignment if he/she hands them over to the first carrier but cannot identify them with appropriate markings, shipping documents, etc.?
\\\\
This problem can be formulated as a question concerning the derivability of the following conditional: $(\bigwedge UN \wedge D0.1 \wedge D1 \wedge \neg D1.1) \rightarrow Ob(D1.2)$. The answer to Query 3 is positive if and only if the conditional is provable in \textbf{DL}$^*$.

\paragraph*{Query 4.}
Is the buyer allowed to claim for damage in case the contract requires the seller to take care of the carriage and the seller neither effects a carriage insurance nor provides the buyer information to effect such an insurance?
\\\\
This problem can be expressed as a question concerning the derivability of the following conditional in \textbf{DL}$^*$: $(\bigwedge UN \wedge D0.1 \wedge \neg D1.4 \wedge \neg D1.5)\rightarrow Pm(E2)$. The answer is positive if and only if the conditional is provable.

\paragraph*{Query 5.}
Is the buyer allowed to exercise rights in case the contract makes reference to goods to be produced at a particular place and the seller disposes them at his/her address of business?
\\\\
This problem can be expressed as a question concerning the derivability of the following conditional in \textbf{DL}$^*$: $(\bigwedge UN \wedge D0.2 \wedge D3)\rightarrow Pm(E1)$.
The answer is positive if and only if the conditional is provable.

\paragraph*{Query 6.}
Is the buyer allowed to exercise rights in case the goods are not delivered according to the procedure described in the contract?
\\\\
This problem can be expressed as a question concerning the derivability of the following conditional in \textbf{DL}$^*$: $(\bigwedge UN \wedge \neg D)\rightarrow Pm(E1)$.
The answer is positive if and only if the conditional is provable.




\subsection{Using MleanCoP}
\label{sec:mleancop}

In order to automate the answering of such questions, we need an efficient implementation of a proof-calculus for \textbf{DL}$^*$.
Several proof-calculi and implementations exist for multimodal versions of \textbf{SDL} and, according to a result in \cite{deBoer&al.12} mentioned in
section \ref{Sec 2}, the logic \textbf{DL}$^*$ turns out to be a bimodal version of \textbf{SDL} exploiting a non-primitive modal operator, $Ought^*$, to express obligations.
This fact allows us to use
standard theorem-provers for normal multimodal logic in order to check the derivability of a formula in \textbf{DL}$^*$.
We also want to mention that a proof-calculus for the related system \textbf{DL} is developed in \cite{kem}; however, such calculus has no implementation and, as we argued in section \ref{Sec 2}, there are many theoretical reasons to prefer \DLS over \textbf{DL} in order to represent normative scenarios involving contrary-to-duty reasoning.

Among the various systems implementing proof-calculi for normal multimodal logic two are prominent. The first, MleanCoP\footnote{\url{http://www.leancop.de/mleancop/}} (\citealt{mleancop}), is a native theorem-prover for various systems of normal modal logic, among which multimodal versions of \DLS. The second, Leo3 (\citealt{leo3}), is a theorem-prover for higher-order logic that can be exploited by translating formulas of a modal language into formulas of a higher-order non-modal language.
Each method has its benefits and limitations. Here we choose to employ MleanCoP, which provides an efficient method and builds proofs
directly within the modal language, so no backward translation from higher-order logic is needed. 
%
In order to ask MleanCoP the queries in section \ref{Sec 3}, we have to translate them into a format which MleanCoP
can understand. Currently, MleanCoP supports two different formats for codifying logical languages, the general format TPTP \footnote{\url{http://www.tptp.org/}}
and its own specific one. We will use MleanCoP's own format, since it is more concise.

\begin{definition}[MleanCoP's syntax]
An MleanCoP problem is a predicate of the form $f(G).$ where $G$ stands for an arbitrary formula.
Formulas are constructed from atoms, whose name must start with a lowercase letter, and the following operators:
\begin{itemize}
  \item The standard propositional operators \verb+'~'+ (negation), \verb+';'+ (disjunction), \verb+','+ (conjunction),
    \verb+'=>'+ (implication) and \verb+'<=>'+ (equivalence).
  \item The modal box operators \verb+'# 1^d: G'+ (\verb+G+ holds in all ideal worlds) and \verb+'# 2^d: G'+ (\verb+G+ holds in all awful worlds).
  \item The modal diamond operators \verb+'* 1^d: G'+ (\verb+G+ holds in some ideal worlds) and \verb+'* 2^d: G'+ (\verb+G+ holds in some awful worlds).
\end{itemize}
\end{definition}

A comparison between our formalization of the Convention in section \ref{sec:form} and MleanCop's syntax points out that the encoding of queries in MleanCoP is very laborious. In order to make the use of this theorem-prover
more accessible, we developed a program capable of executing queries written in our encoding. This program translates
the encoding into the syntax of MLeanCop and then executes the prover. The programs discussed in this paper as well as the latest
version of MLeanCop can be downloaded from Zenodo.
\footnote{\url{https://zenodo.org/record/1450890\#.W7oKGS1fgWM}}

\begin{definition}[The program syntax]
The input to the program consists of a predicate of the form \verb+([ls],F)+ where \verb+[ls]+ is a list of comma separated assumptions
and $F$ is the goal we try to prove. Atom names must respect the syntax of MLeanCop. Both assumptions and goal are constructed using the following operators, which imitate those in Definition \ref{df:syntax}.
\begin{itemize}
  \item The standard propositional operators as defined by MLeanCop;
  \item The ideality operator \verb+Id(F)+ and the awfulness operator \verb+Aw(F)+ applied to a formula \verb+F+;
  \item The obligation operator \verb+Ob(F)+ and the permission operator \verb+Pm(F)+ applied to a formula \verb+F+;
  \item The conditional obligation \verb+NO(F,G)+ and the conditional permission \verb+NP(F,G)+ applied to a pair of formulas \verb+F+ and \verb+G+.
\end{itemize}
In addition, in order to simplify the process of asking questions, we have added a constant \verb+un+ which codifies the conjunction of the formulas (1)-(18) in Definition \ref{df:un}, namely $\bigwedge UN$.
\end{definition}

MleanCoP is written in Prolog and an installation of one of the supported distributions of Prolog is required to run it. Our program
comes bundled with MleanCoP version 1.3 and requires Ruby version 2.5.1 and the Ruby's gem 'bundler'
version 1.16.2. We tested our program on Debian 9 using SWI-ptolog version 7.2.3.
Let $A$ be a formula in the language of \textbf{DL}$^*$ given as an input: the Ruby program translates $A$ into the format compatible with MleanCoP and executes the latter. The possible answers are
`Theorem' and `Non-theorem'. The answer `Theorem' means that the formula given as an input is provable in \DLS; the answer `Non-theorem' means that the formula given as an input is not provable in \textbf{DL}$^*$.

In addition, an answer of Theorem is accompanied by a proof of derivability in the connection calculus employed by MLeanCop (\citealt{Bibel83}).
As an example of the procedure described so far, we show how one can formulate Query 1 from section \ref{sec:is}.
In order to adhere to the syntax of MLeanCop, we use lower case letters to encode the propositional constants from
Definition \ref{df:un} and we simplify a bit the names of formulas (e.g., $D0.1$ becomes $d01$).

\begin{lstlisting}
ruby prove1.rb "([un,d01,d3],((~ d);((~ p);(~ g))))"
problem is a modal (multi/const) Theorem
Start of proof for problem
...
End of proof for problem

\end{lstlisting}

Once the above is executed, MleanCoP states that this is a Theorem and returns a proof. The commands to execute the program on
the remaining queries can be found in Figure \ref{fig:answers}.
The method just described is efficient. The total time of running all involved programs on each query takes about a second \footnote{The program was executed on a laptop with an Intel Core i7-5600U processor.}.

\subsection{Legal applications of automated reasoning}

A framework for normative reasoning as the one discussed in this paper can be used for various
applications. In this section we discuss two of them.

The first application is a tool for helping a subject to navigate a legal text, such as the Convention. Such a tool can make a legal text more accessible to non-experts. A subject only needs to be able to install the required software and to build queries; however, a web application and a simple user interface which takes
sentences as inputs and translates them into the intended expressions to be checked can make make the whole process more human-friendly. 

We now show the possible executions of our program, which correspond to the six queries of section \ref{sec:is}. Each line in the table in Figure \ref{fig:answers} displays one of such executions. Each
execution requires a set of current facts and an actual or ideal obligation. The user might choose this information from a pre-defined
set of options (by using a drop-box, for example). The remaining part of each line shows the query being sent to the prover and its response.

\begin{figure}
  \tiny
  \begin{tabular}{|p{2.5cm}|p{1.6cm}|p{1.2cm}||p{5.5cm}|p{0.8cm}|}
Facts & Goal & Type & Command & Answer \\
\hline
$D0.1, D3$ & $\neg D \vee \neg P\vee\neg G$ & Violation & \verb+([un,d01,d3],((~ d);((~ p);(~ g))))+ & Yes\\
$\neg D0.1, \neg D0.2, \neg D3$ & $\neg D\vee\neg P\vee\neg G$ & Violation & \verb+([un,(~ d01),(~ d02),(~ d3)],((~ d);((~ p);(~ g))))+ & Yes\\
$D0.1, D1, \neg D1.1$ & $D1.2$ & Obligation & \verb+([un, d01,d1,(~ d11)],(Ob d12))+ & Yes\\
$D0.1, \neg D1.4, \neg D1.5$ & $E1$ & Permitted & \verb+([un,d01,(~ d14),(~ d15)],(Pm e1))+ & Yes\\
$D0.2, D3$ & $E2$ & Permitted & \verb+([un,d02,d3],(Pm e2))+ & Yes\\
$\neg D$ & $E2$ & Permitted & \verb+([un,(~ d)],(Pm e2))+ & Yes\\
 & $G$ & Ideal & \verb+([un],(Id (Ob g))+ & Yes \\
  \end{tabular}
  \label{fig:answers}
  \caption{A tool for helping buyers navigate the UN convention}
\end{figure}

Another application might be as a courtroom decision support system. Decision support systems are widely used in many
fields, such as management (\citealt{mng}), medicine (\citealt{med}) and civil engineering (\citealt{eng}). Such systems help subjects in making informative decisions. 
These systems do not
replace a human in decision-making, they simply rather exclude from the set of possible decisions those which do not comply with a given
reality.

Decision-supporting systems in the courtroom, which are intended in helping judges make the right decisions, are much less common.
In fact, to the best of knowledge of the authors, just a handful of such systems exist.  Winjuris\footnote{\url{http://softpert.com/legal/court-management/winjuris}}
and Forecourt\footnote{\url{https://www.rsi.com/products/forecourt/}} are management systems rather
than decision-supporting systems. More interesting examples are an Israeli system for the evaluation of criminal records (\citealt{offender}),
the Australian `Split Up' system (\citealt{splitup}), which assists in property splitting during divorce trials, and a case-based reasoning
system for a `virtual courtroom' (\citealt{virtual}).

One can apply the framework presented in this paper in order to display all possible normative outcomes for a given
set of facts: A judge can feed the system with a set of known facts and a set of unknown parameters, as well as with a set of possible
interesting outcomes. One option is then to display the information as a graph showing all possibilities which are compatible with the relevant body of laws.
Using this graph, a judge can more easily navigate the set of all possible decisions and can avoid making decisions which are not
supported by the facts and norms of the case.

\begin{sidewaysfigure}
  \includegraphics[scale=0.3]{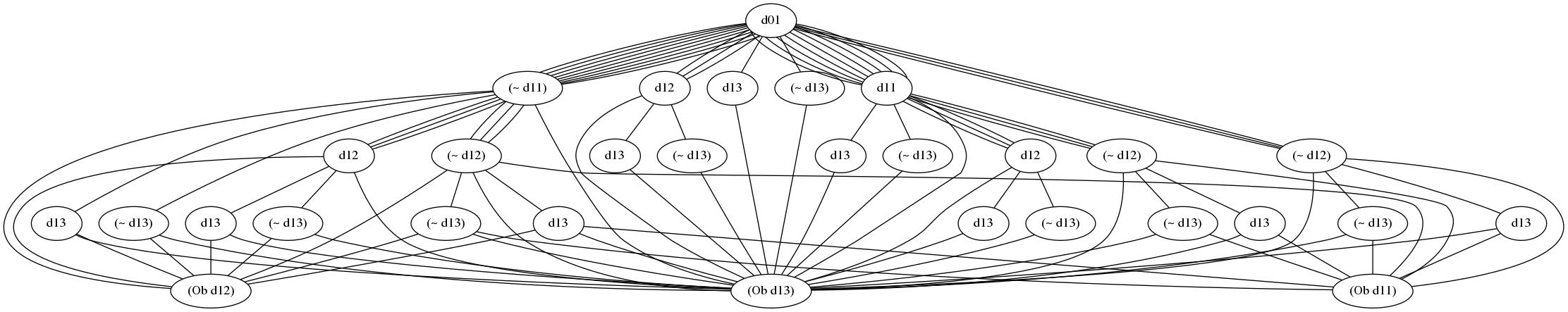}
    \caption{A graph representing all possible normative paths for a certain case.}
    \label{fig:tree}
\end{sidewaysfigure}

We developed a second program for generating such graphs.
Given the input just mentioned, the program executes the script from section  \ref{sec:mleancop}
on all possible combinations and builds the corresponding graph.
Figure \ref{fig:tree} shows the execution of the program given the
facts $\{D0.1\}$, unknown parameters $\{D1.1, D1.2, D1.3\}$ and possible outcomes $\{Ob(D1.1), Ob(D1.2), Ob(D1.3)\}$.

To obtain the above image in \verb+dot+ format,\footnote{\url{https://
www.graphviz.org/doc/info/lang.html}} a user should execute the following command:

\begin{lstlisting}
> ruby tree.rb "d01" "d11,d12,d13"
  "(Ob d11),(Ob d12),(Ob d13)" > mytree.dot
\end{lstlisting}

\section{Final remarks}
In this article we provided an automatic procedure for reasoning on legal texts. We relied on a logic of normative ideality and sub-ideality called \textbf{DL}$^*$, presented in \cite{deBoer&al.12}, and focused on problems that can be represented in terms of a Normative Detachment Structure with Ideal Conditions ($NDSIC$). Our theoretical framework provides also some ground for reflection on the problem of the formal rendering of contrary-to-duty scenarios. Going back to the specific case of Chisholm's paradox, we proposed to formalize sentences (1)-(4) in section \ref{Sec 2} as follows (here we employ again the notation used in the literature):
\begin{itemize}
\item[(1a')] $O (Ought^* (P))$;
\item[(2a'')] $(P\rightarrow Ought^*(Q))\wedge O(Ought^*(P)\rightarrow Ought^*(Q))$;
\item[(3a'')] $(\neg P\rightarrow Ought^*(\neg Q))\wedge O(Ought^*(\neg P)\rightarrow Ought^*(\neg Q))$
\item[(4a)] $\neg P$.
\end{itemize}
Formula (1a') is suggested by \cite{J&P85} and wants to stress that Jane ought to help her neighbors in normatively ideal circumstances (though, not necessarily in the actual circumstance). Formula (4a) is the obvious way of rendering the fact that Jane does not help her neighbors. Formulas (2a'') and (3a'') are a novelty to represent conditional obligations. They were obtained by putting together some intuitions in \cite{deBoer&al.12} and the possibility of building a chain of ideal obligations; in particular, since Jane ideally ought to help her neighbors, then she ideally ought to tell them. While being more sophisticated than alternative logical renderings of Chisholm's scenario, this solution has several advantages which can be highlighted by making reference to some discussion in \cite{Carmo&Jones02}. Indeed, in the latter work the authors list \textit{eight criteria} that should be met by a logical representation of contrary-to-duty scenarios:
\begin{itemize}
\item[(i)] consistency;
\item[(ii)] logical independence of the members;
\item[(iii)] applicability to (at least apparently) timeless and actionless contrary-to-duty examples;
\item[(iv)] analogous logical structures for conditional sentences;
\item[(v)] capacity to derive actual obligations;
\item[(vi)] capacity to derive ideal obligations;
\item[(vii)] capacity to represent the fact that a violation of an obligation has occurred;
\item[(viii)] capacity to avoid the pragmatic oddity.
\end{itemize}

We have already shown that our formalization meets the requirement (viii) and the reader can easily check that (i) and (ii) are met as well. Criterion (iii) is also satisfied, since this formalization is not time-dependent (the language of \textbf{DL}$^*$ cannot make temporal distinctions). The fact that conditional obligations always have the same logical rendering is evident, so we can mark with a tick criterion (iv) too. Criteria (v), (vi) and (vii) require further analysis. Actual obligations are those of kind $Ought^*(A)$ and in our representation of Chisholm's scenario one can surely infer $Ought^*(\neg Q)$: Jane actually ought not to tell her neighbors that she is coming (since she decided not to help them). Ideal obligations are those of kind $O(Ought^*(A))$ and in our representation of Chisholm's scenario one can infer $O(Ought^*(Q))$ from $O(Ought^*(P))$ and $O(Ought^*(P)\rightarrow Ought^*(Q))$, since the schema $O(A\rightarrow B)\rightarrow (OA\rightarrow OB)$ is provable in \textbf{DL}$^*$: Jane ideally ought to tell her neighbors that she is coming because she ideally ought to help them. Thus, both (v) and (vi) are met. Finally, concerning criterion (vii), we can say that a violation of $A$ occurs when $A$ ideally ought to be the case, it is currently not the case but it could have been the case in normatively ideal circumstances. Now, in our scenario we have both $O(Ought^*(A))$ and $\neg A$ as premises; let us take these formulas as true in the actual world, call it $w_a$. From $O(Ought^*(A))$ one can infer $OPA$, which means that all worlds which are normatively ideal with respect to  $w_a$ have access to a (normatively ideal) world in which $A$ is the case. We can paraphrase this as follows: in all ideal circumstances with respect to $w_a$ it is possible to bring about $A$, despite $A$ not being the case in $w_a$. This is the way in which our approach can represent the fact that a violation of an (ideal) obligation occurred.

A last comment concerns our choice to use a simple logic for the normative reasoning discussed in this paper. In the introduction, we
have mentioned that this choice is primarily motivated by complexity issues. Indeed, already our simple encoding of parts of the Convention
involves hundreds of terms. Nevertheless, MLeanCop manages to answer each query in about a second. At the same time, there are
many interesting questions which are beyond the tools mentioned in this paper. For example, abductive reasoning (\citealt{abduction}) can provide explanations
for the invalidity of queries. By choosing a normal modal logic such as \DLS, we ensure the possibility of using such results in the future.


\end{document}